\documentclass{article}

\usepackage[nonatbib,final]{ml4eng_2020}

\usepackage[utf8]{inputenc} 
\usepackage[T1]{fontenc}    
\usepackage{hyperref}       
\usepackage{url}            
\usepackage{booktabs}       
\usepackage{amsfonts}       
\usepackage{nicefrac}       
\usepackage{microtype}      
\usepackage{biblatex}
\usepackage{graphicx}
\usepackage{subcaption}
\usepackage{amsmath}
\usepackage{booktabs}
\usepackage{cleveref}
\usepackage{booktabs}

\title{Multi-Loss Sub-Ensembles for Accurate Classification with Uncertainty Estimation}

\author{%
  Omer Achrack \\
  Intel (EGI)\\
  \texttt{omer.achrack@intel.com} \\
   \And
  Raizy Kellerman \\
 Intel (Mobileye)\\
  \texttt{raizy.kellerman@intel.com} \\
    \And
  Ouriel Barzilay \\
 DermaDetect Ltd. / Intel (Mobileye)\\
  \texttt{ouriel@dermadetect.com} \\
}

\begin{document}

\maketitle

\begin{abstract}
Deep neural networks (DNNs) have made a revolution in numerous fields during the last decade. However, in tasks with high safety requirements, such as medical or autonomous driving applications, providing an assessment of the model's reliability can be vital. Uncertainty estimation for DNNs has been addressed using Bayesian methods, providing mathematically founded models for reliability assessment. These model are computationally expensive and generally impractical for many real-time use cases. Recently, non-Bayesian methods were proposed to tackle uncertainty estimation more efficiently.
We propose an efficient method for uncertainty estimation in DNNs achieving high accuracy. We simulate the notion of multi-task learning on single-task problems by producing parallel predictions from similar models differing by their loss. This multi-loss approach allows one-phase training for single-task learning with uncertainty estimation. We keep our inference time relatively low by leveraging the advantage proposed by the Deep Sub-Ensembles method.
The novelty of this work resides in the proposed accurate variational inference with a simple and convenient training procedure, while remaining competitive in terms of computational time.
We conduct experiments on SVHN, CIFAR10, CIFAR100 as well as ImageNet using different architectures. Our results show improved accuracy on the classification task and competitive results on several uncertainty measures.
\end{abstract}

\section{Introduction}
Deep neural networks have achieved impressive results in various A.I. areas such as computer vision, natural language processing and audio analysis. The common approach for classification in supervised learning is to employ softmax activation on the logits vector. While being efficient and intuitive, it lacks the ability to provide any kind of uncertainty measure that would yield an estimation of what the model does not know (e.g. a lack of expertise on a given input) \cite{kendall2017uncertainties,relaxed_SM,TS_MC_drop}. These models generally tend to be over-confident as mentioned in\cite{relaxed_SM,temperature_scaling}. When applied to life-critical tasks, such as autonomous driving and medical diagnosis, the results might be devastating.
In such use cases, the model should have the essential ability to determine what it knows and what it does not. Ideally, this ability should be reflected by a proper confidence measure associated to the model prediction, likely to prevent rare, but costly events of over-confident mispredictions. Bayesian techniques \cite{prior_1_mackay1992practical,prior_2_mackay1992bayesian,prior_3_hinton1993keeping} offer a theoretically grounded approach to infer uncertainty from a model, but their computational resources consumption often make them impractical. Several non-Bayesian methods have been proposed for providing uncertainty measures \cite{gal2015dropout,deep_ensebles}. While being more computationally attractive than Bayesian methods, these methods still require multiple forward passes at inference time to obtain a reliable uncertainty measure. Recently, the Deep Sub-Ensembles (DSE) \cite{valdenegro2019deep} method proposes a simplification of the Deep Ensembles (DE) \cite{deep_ensebles} method by sharing a common network trunk and composing different heads, called \textit{sub-ensembles}.\newline 
 We propose an enhancement to the DSE method by associating a different loss function to each sub-ensemble. As a result, training can be performed in an end-to-end manner, instead of separately for each network branch, saving computational resources. Furthermore, our method improves the model's accuracy, and offers competitive uncertainty estimation results, considering the attractive run time and convenient training procedure, as shown in  Section 4.\newline
We organize the rest of the paper as follows: Section 2 briefly overviews the recent methods addressing model confidence. Section 3 describes the details of the proposed method. In Section 4, we show our experimental setup and results, which are discussed in Section 5. We conclude and propose directions for future work in Section 6.

\section{Related work}
\paragraph{Types of uncertainty}  
In Bayesian learning, there are two main types of uncertainty one can model \cite{der2009aleatory}. Aleatoric uncertainty (from Greek \textit{alea}, “rolling a dice”), is a \textbf{data property} capturing the noise inherent to observations and thus, can't be reduced even if more data is collected. On the other hand, Epistemic uncertainty (from Greek \textit{episteme}, “knowledge”) is a \textbf{model property} and can be reduced if more data is collected. This uncertainty measure stems from the training process: formally, it estimates a posterior distribution over the weights $w$ given the data set $\mathcal{D}$, i.e. $\mathcal{P}(w|\mathcal{D})$, in contrast to $\mathcal{P}(y|x,w)$ where $x$ and $y$ are respectively the network's input and output. The epistemic uncertainty measure could also be interpreted as the result of an insufficiency in the training data at certain regions of the input space, \cite{kendall2017uncertainties}, affecting in turn prediction reliability. In this work, we address epistemic uncertainty estimation.

\paragraph{Non-Bayesian methods for uncertainty estimation}
Many recent studies seek to provide uncertainty estimation using non-Bayesian methods. MC-dropout \cite{gal2015dropout} approximates at inference time the posterior distribution of the weights by incorporating a dropout \cite{hinton2012improving_dropout,ghiasi2018dropblock,chen2020dropcluster} mechanism. By performing the forward pass of the network several times, the method generates different predictions induced by randomized dropouts on the activations, to model the epistemic uncertainty. 
In Deep Ensembles \cite{deep_ensebles}, several differently-initialized models are trained with random shuffling of the training data to obtain model variations given a training set. In this way, each model is treated as a sample from $\mathcal{P}(w|\mathcal{D})$. At inference time, the method averages probability prediction vectors over the different models, and quantifies the uncertainty by means of the averaged probability's entropy. While outperforming MC-dropout \cite{valdenegro2019deep}, DE is expensive in terms of memory and compute resources for both training and inference. Tagasovska et al. \cite{tagasovska2019single} propose a single model estimating both types of uncertainty, using quantile regression for aleatoric uncertainty and out-of-distribution detection, by using a set of different linear classifiers for epistemic uncertainty.
In \cite{ahuja2019probabilistic}, the authors propose to model the deep feature space at different intermediate layers of the network using multivariate Gaussian or Gaussian Mixture Models (GMM) for enabling the model to detect out-of-distribution samples, via Mahalanobis distance \cite{chandra1936generalised_maha}. This work, by performing a separate modeling for each class, extends \cite{lee2018simple} where the feature distributions are estimated for the entire data set as a whole. The post-training modeling required by these methods is relatively expensive since it involves computing multivariate Gaussians or GMMs.
Jiang et al. \cite{jiang2018trust} train KNN classifiers using the intermediate feature space for each class at selected layers. The KNN scores are then used to determine the prediction's reliability. 
Kendall et al. \cite{kendall2017uncertainties} combine the modeling of both uncertainty measures: the aleatoric uncertainty estimator is trained by a dedicated branch of the network, while the epistemic uncertainty is modeled using MC-dropout.
Another recent approach \cite{sensoy2018evidential} models the posterior probability of the predicted class by directly learning the alpha coefficients of the Dirichlet distribution \cite{samuel2004continuous}. The uncertainty was quantified as the ratio of the number of classes to the alpha coefficients sum.
Corbiere et al. \cite{corbiere2019addressing} train a network aiming at predicting the confidence, based, in an innovative manner, on the true class probability (TCP), rather than on the maximum class probability like softmax. Their network includes a separate branch trained to regress the TCP with $L2$ norm as a dedicated loss function for that branch.
In the widely-used Temperature Scaling (TS) \cite{temperature_scaling} method, a TS parameter is used to normalize the final logits vector for control on the "peakiness" of the model's feature vector before the final softmax activation layer. The parameter can either be manually tuned using a dedicated validation set, or being part of the predicted values as in \cite{relaxed_SM}. According to \cite{TS_MC_drop} combining TS \cite{temperature_scaling} with MC-dropout, the TS method significantly reduces the expected calibration error (ECE) on several data sets. 
Post-Hoc Calibration \cite{kull2017beyond} aims at learning a re-calibration matrix for reducing the ECE on the predictions produced by the softmax activation \cite{Beyond_temperature_scaling}. This is done under the assumption that the re-calibration matrix coefficients are sampled from a Dirichlet distribution modeled by a fully connected layer.
\paragraph{Bayesian methods for uncertainty estimation}
The method proposed in \cite{Weight_uncertainty} models the epistemic uncertainty by directly learning $\mathcal{P}(w|\mathcal{D})$. At inference time, the model parameters are sampled from the learned posterior distribution to produce multiple predictions and perform variational inference. An extension of this method by \cite{Bayesian_layers} provides an open-source implementation with additional Bayesian RNN and LSTM layers. However, these approaches use the relaxed assumption of diagonal-dominant covariance matrices for the posterior distribution of the weights in each layer.
Another innovative approach \cite{VADAM} proposes to perturb Adam \cite{ADAM_optim} gradients during training and show improvements on the uncertainty estimation performed with MC-dropout.
The Stochastic Weight Averaging (SWA) method \cite{SWA} proposes to train a model from scratch until convergence, and in a next step, to train it further for a smaller number of epochs. At each of those additional epochs, a snapshot of the model weights is collected. These sets of weights are then averaged to produce a model approximating Deep Ensembles. While significantly improving run-time requirements, this method achieves impressive results on several data sets, as mentioned by \cite{SWAG}, which recently extended SWA by approximating, from the learned models, the posterior distribution over the weights. The authors also relax the diagonal-dominant assumption on the covariance matrix and use instead low rank covariance matrices for estimating the posterior distributions over the weights of the whole network.
\section{Method}
\label{seq:method}
\begin{figure*}
    \begin{center}
    \begin{subfigure}[b]{0.45\textwidth}  \includegraphics[width=0.9\linewidth]{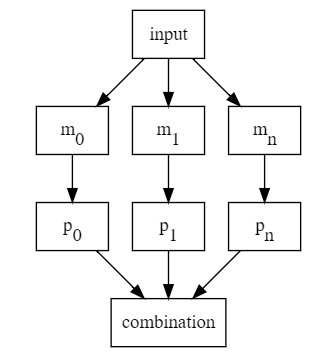}
        \caption{Deep Ensembles \cite{deep_ensebles}}
    \end{subfigure}
    \begin{subfigure}[b]{0.45\textwidth}
        \includegraphics[width=0.9\linewidth]{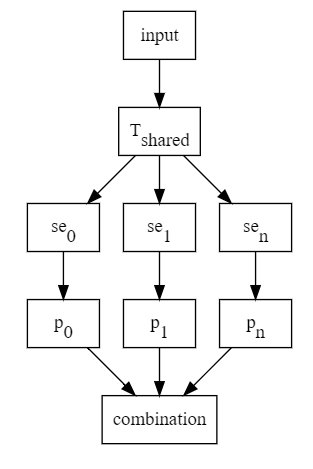}
        \caption{Deep Sub-Ensembles \cite{valdenegro2019deep}}
    \end{subfigure}
        \caption{Schematic comparison of Deep ensembles \cite{deep_ensebles} vs. Deep Sub-Ensembles \cite{valdenegro2019deep} (as depicted in \cite{valdenegro2019deep}). Deep ensembles do not share any layer, while deep sub-ensembles train the trunk only once and share that trunk across sub-ensembles.}
        \label{fig:ensembles_vs_subensembles}
    \end{center}
\end{figure*}

\begin{figure*}[ht]
    \centering
    \begin{subfigure}[t]{0.43\textwidth}
    \includegraphics[width=0.9\textwidth,keepaspectratio]{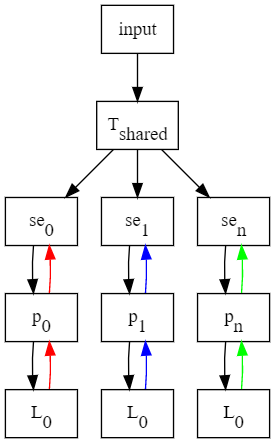}
    \captionsetup{width=0.8\textwidth}
    \caption{Training procedure of the DSE approach \cite{valdenegro2019deep}: each branch $se_i$ is trained separately using the same loss function.}
    \end{subfigure}
    \begin{subfigure}[t]{0.43\textwidth}
    \includegraphics[width=0.9\textwidth,keepaspectratio]{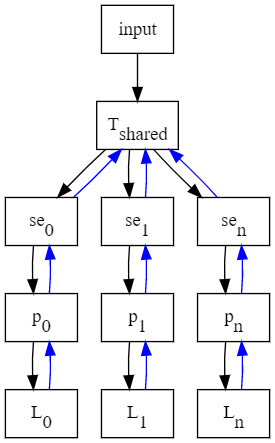}
    \captionsetup{width=0.8\textwidth}
    \caption[width=0.8\linewidth]{The proposed approach: the training is end-to-end. Each branch is optimized with its own loss function.}
    \end{subfigure}
    \caption{Schematic view of the training procedures used in our approach and in \cite{valdenegro2019deep}. Black arrows represent the forward pass and colored arrows depict the back-propagation paths in the network. Different colors denote different training phases.}
    \label{fig:our_train_approach_vs_dse_app}
\end{figure*}

\paragraph{Deep Ensembles and Deep Sub-Ensembles}
In Deep Ensembles \cite{deep_ensebles}, $M$ models are separately trained on a data set $D=\{x_i,y_i\}$ via Maximum Likelihood Estimation (MLE) optimization, implemented as Stochastic Gradient Descent (SGD) or any of its variants \cite{ruder2016overview_SGD}. Once the training of the different models has been performed, each classifier $i$ provides a prediction vector $\hat{y_i}$ representing the probability distribution over the defined classes $\{1 \dots C\}$. These $M$ distributions are eventually averaged to compose the final prediction $\hat{y}$  (see \autoref{eqn1} and e.g. \cite{TS_MC_drop,valdenegro2019deep}).
\begin{equation}
    \hat{y} =  \frac{1}{M} \sum_{i=1}^{M}{\hat{y_i}}    
    \label{eqn1}
\end{equation}
Alternatively, in DSE \cite{valdenegro2019deep}, the network shares a trunk and sequentially train each of the sub-ensembles. Since our approach is similar to the DSE method, we depict the difference in the network architectures of DE and DSE in \autoref{fig:ensembles_vs_subensembles}.
\paragraph{Multiple views on the same task}
We propose to train the whole network from scratch in a single phase (in contrast to DE or SWA) and in an end-to-end manner by combining the advantages of the paradigms of variational inference (VI) and multi-task learning. 
In multi-task learning \cite{ren2015faster,he2017mask_rcnn}, a single network, with shared layers and a dedicated head for each task, is trained in an end-to-end manner to perform simultaneously different tasks on the same input data. This usually improves the results on \emph{each task} separately. Inspired by this notion, we propose to train, on a \emph{single task}, a network with DSE-like architecture where each of the cloned heads is optimized using a different loss function. The rationale behind that lays on the conjecture that each head could benefit from the shared knowledge obtained from other "experts" in the same field.
A comparison between the training procedures of our approach and DSE \cite{valdenegro2019deep}, is demonstrated in \autoref{fig:our_train_approach_vs_dse_app}.
We next review the different loss functions that were utilized for assessing the proposed approach. Sensoy et al. \cite{sensoy2018evidential} propose a loss allowing quantification of the classification uncertainty using Dirichlet distribution. Relaxed softmax \cite{relaxed_SM} predicts an additional parameter used for normalizing the other features, similarly to \cite{temperature_scaling}, for dealing with miscalibrated softmax probability vectors. Xu et al. \cite{l_dmi_loss} address the classification objective from a different perspective: their classifier is trained using an information-theoretic-based loss aiming at overcoming data with noisy labels. With the same goal of robustness against noisy labels, Amid et al. \cite{amid2019robust_bi_tempered} propose to couple a novel activation function, called Tempered softmax, with an adequate loss\footnote{As accuracy was not improved using Tempered softmax loss, the results on these experiments were omitted from \autoref{seq:exp}}. 
$L1$ and $L2$ norms have commonly been used as loss functions for classification and regression problems. These loss functions are distance-based and very similar one to another or to the cross-entropy loss already used in the relaxed softmax head. However and as these norms provided notable improvements, we included $L1$ and $L2$ to the set of selected loss functions.   
After convergence of the training, we perform variational inference (VI) as described in \autoref{eqn1} and similarly to DSE \cite{valdenegro2019deep}.

\paragraph{Uncertainty estimation measures} Estimation of the model uncertainty might be done in several manners. One option would be to use the normalized entropy of the averaged probability vector as in \cite{TS_MC_drop}.
\begin{equation}
    u = \frac{1}{\log(C)} \sum_{i=1}^{C} \overline{y}_i\log(\overline{y}_i)
    \label{eqn2}
\end{equation}
where $\overline{y}$ indicates the \textbf{averaged} probability vector and $C$ is the number of classes.
Alternatively, the normalized entropy could be averaged for each prediction vector $y_i$ produced by the sub-ensemble $se_i$.
\begin{equation}
    u = \frac{1}{M}\frac{1}{\log(C)} \sum_{j=1}^{M} \sum_{i=1}^{C}  \hat{y}_i\log(\hat{y}_i)
    \label{eqn3}
\end{equation}
where $\hat{y}$ indicates a \textbf{single} probability vector.
We also introduce a metric for uncertainty measure on variational inference setups, estimated as the variance of the predicted class $j$ across the different predictions $\hat{y}_{j_i}$, formally:
\begin{equation}
\begin{aligned}
    u =& \sum_{i=1}^{M}{(\hat{y}_{j_i} -\mu)^2}\\
    j&=\underset{\hat{y}}{\arg\max}\\
    \mu &= \frac{1}{M}\sum_{i=1}^{M}{\hat{y}_{j_i}} \\
\end{aligned}
\label{eqn4}
\end{equation}

\section{Experiments}
\label{seq:exp}
\paragraph{Experiment design intent} In order to make the approach available for practical use cases, we employ a modest number of heads, even for scalable methods. We employ the loss functions mentioned in \autoref{seq:method} and compare the obtained results with MC-dropout \cite{gal2015dropout}, Deep Ensembles \cite{deep_ensebles}, Deep Sub-Ensembles\cite{valdenegro2019deep} as well as Stochastic Weight Averaging (SWA) \cite{SWA}.
\paragraph{Experimental setup} We use state-of-the-art neural network architectures in order to evaluate our approach under various conditions. From the family of residual architectures, we choose  MobileNet-v2 \cite{mobilenetv2}, ResNet50\cite{he2016deep_resnet} and Xception\cite{chollet2017xception}. Additionally, we include VGG-16 \cite{VGG}. Each of these architectures was trained and tested using Keras\cite{chollet2015keras} on CIFAR10 \cite{cifar10}, CIFAR100\cite{cifar100} and ImageNet \cite{imagenet_cvpr09} data sets respectively, for residual architectures . VGG-16 was trained and tested on SVHN \cite{SVHN}. In all experiments\footnote{The goal is to examine our method and not necessarily find out the best training procedure, hence we chose as simple as possible a training procedure for a uniform and fair comparison}, the learning rate was set as $1e^{-3}$. We calculate the accuracy and Brier score \cite{brier1950verification} 
defined as $BS = \frac{1}{C}\sum_{i=1}^{C}{(t^{*}_{i} - p(y = i|x))^2}$ where $t^{*}_{i}$ = 1 if $i = y^{*}$, and 0 otherwise. Additionally, we compute the expected calibration error (ECE) \cite{niculescu2005predicting_ECE1,guo2017calibration_ECE2}. The models with best accuracy on the test set among 300 epochs of training were selected for evaluation. In the DSE \cite{valdenegro2019deep} method and ours, the split point between the trunk and the sub-ensembles was defined with regards to the internal structure of the selected architecture. Specifically, the split point was set for MobileNet-v2, at the last two inverted residual blocks (with 160 and 320 channels, respectively);
for ResNet50, at the last residual block (with 2048 channels);
for VGG16, at the last 3 convolution blocks (with 128, 256 and 512 channels respectively).
For Deep Ensembles, we trained the vanilla model, with no dropout nor any other regularization. We chose relaxed softmax \cite{relaxed_SM} as a base loss for experiments with MobileNet-v2, VGG16 and Xception, while for ResNet50, we chose the standard softmax for the sake of experimental diversity. After the training converged, we additionally trained the models for 30 epochs, and selected the best $N$ models to compose the ensemble. Those models were also used to calculate the Stochastic Weight Averaging \cite{SWA} model, with no additional fine-tuning. For VGG16 and MobileNet-v2, $N=4$; for ResNet50, $N=3$; and for Xception $N=2$.
In MC-dropout, the dropout probability was set as $0.5$ for all architectures. The dropout layer was placed after each block depending on the internal structure of the architecture. For the sake of fair evaluation, we set the various key parameters in the most similar fashion for the different methods. Specifically, $N$ was shared across methods given a pair of architecture/data-set, i.e. $N$ ensembles, $N$ heads for DSE, and $N$ stochastic forward passes for MC-dropout. Implementation details and results are summarized in\autoref{tbl:impl_details} and in \autoref{tbl:basic_comp}, respectively.            

\begin{table}[ht]
 \centering
  \caption{Basic accuracy and calibration comparison}
  \label{tbl:basic_comp}
  \begin{tabular}{lllll}
     \toprule
    \multicolumn{2}{c}{}                                                       \\
     Arch/Data-set &Method                  &Accuracy     &ECE     &Brier Score  \\
    \cmidrule(r){1-5}
                     & MC-dropout          &0.965          &0.05       &0.005           \\
                     & DE      &0.963          &0.037         &0.006           \\
VGG16/SVHN           & DSE  &0.962          &0.652      &0.051 \\
                     & SWA                 &0.962 &0.033             &0.006    \\
                     & Ours                &0.966   &0.262           &0.013
           \\
    \cmidrule(r){1-5}
                     & MC-dropout          &0.823        &0.055     &0.025                \\
                     & DE      &0.873        &0.029     &0.019                  \\
MobileNet-v2/CIFAR10 & DSE  &0.861 &0.034            &0.02  \\
                     & SWA                 &0.871 &0.048 &0.022 \\
                     &Ours                &0.875 &0.1 &0.023 \\
    \cmidrule(r){1-5}
                      & MC-dropout         &0.669 &0.029 &0.022 \\
                      & DE     &0.709 &0.074 &0.019 \\
ResNet50/CIFAR100     & DSE &0.609 &0.101 &0.028 \\
                      & SWA                &0.713 &0.021 &0.019 \\
                      & Ours               &0.679 &0.04 &0.022 \\
    \cmidrule(r){1-5}
                     &MC-dropout          &0.581            &0.011       &0.023           \\
                     &DE      &0.801            &0.028       &0.022           \\
Xception/ImageNet    &DSE  &0.802            &0.030       &0.042           \\
                     &SWA                 &0.816            &0.027       &0.022           \\
                     &Ours                &0.832            &0.029       &0.030           \\
    \cmidrule(r){1-5}
  \end{tabular}
\end{table}

\begin{table}[ht]
 \centering
  \caption{Implementation details}
  \label{tbl:impl_details}
  \begin{tabular}{lllll}
     \toprule
    \multicolumn{2}{c}{}                                                       \\
     Arch/Data-set &Method                  &Epochs      &Loss                                                       &Split \\
    \cmidrule(r){1-5}
                     & MC-dropout          &300         &\cite{relaxed_SM}                                                      &-            \\
                     & DE                  &300+30      &\cite{relaxed_SM}                                                      &-          \\
VGG16/SVHN           & DSE                 &300+80x4    &\cite{relaxed_SM}                                                      &Blocks 3-5          \\
                     & SWA                 &300+30      &\cite{relaxed_SM}                                                      &- \\
                     & Ours                &300         &\cite{relaxed_SM}, \cite{sensoy2018evidential}, \cite{l_dmi_loss},MSE  &Blocks 3-5 \\
    \cmidrule(r){1-5}
                     & MC-dropout          &300         &\cite{relaxed_SM}                                                      &-            \\
                     & DE                  &300+30      &\cite{relaxed_SM}                                                      &-          \\
MobileNet-v2/CIFAR10 & DSE                 &300+80x4    &\cite{relaxed_SM}                                                      &Last 2 blocks          \\
                     & SWA                 &300+30      &\cite{relaxed_SM}                                                      &- \\
                     & Ours                &300         &\cite{relaxed_SM}, \cite{sensoy2018evidential}, MAE,MSE  &Last 2 blocks \\
    \cmidrule(r){1-5}                     
                     & MC-dropout          &300         &softmax                                                      &-            \\
                     & DE                  &300+30      &softmax                                                      &-          \\
ResNet50/CIFAR100    & DSE                 &300+80x3    &softmax                                                      &Last block          \\
                     & SWA                 &300+30      &softmax                                                      &- \\
                     & Ours                &300         &softmax,\cite{sensoy2018evidential},MSE                      &Last block \\
    \cmidrule(r){1-5}
                     & MC-dropout          &150         &\cite{relaxed_SM}                                                      &-            \\
                     & DE                  &150+30      &\cite{relaxed_SM}                                                      &-          \\
Xception/ImageNet    & DSE                 &150+50x2    &\cite{relaxed_SM}                                                      &From 7th middle-flow block onward          \\
                     & SWA                 &150+50      &\cite{relaxed_SM}                                                      &- \\
                     & Ours                &150         &\cite{relaxed_SM}, \cite{sensoy2018evidential}                         &From 7th middle-flow block onward \\
    \cmidrule(r){1-5}
  \end{tabular}
\end{table}

\begin{figure*}
    \centering
    \includegraphics[width=0.95\textwidth]{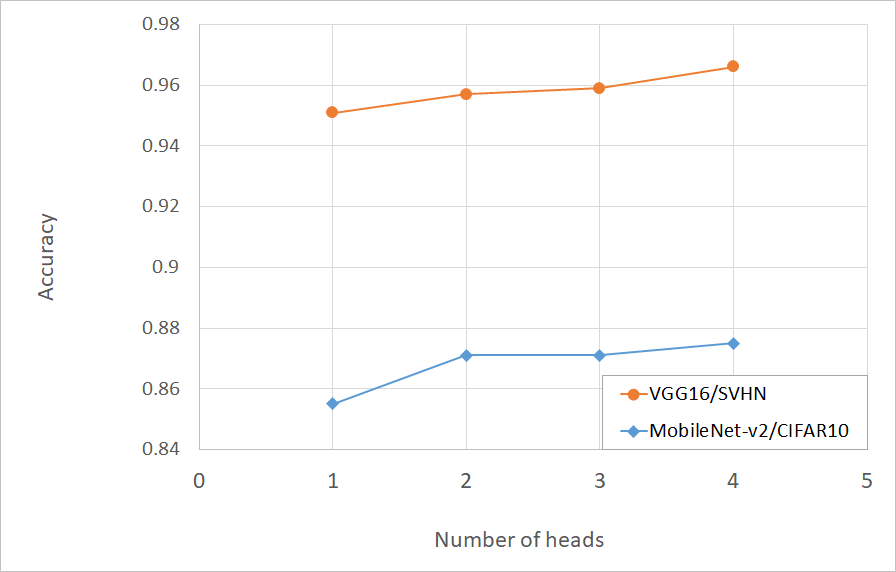}
    \captionsetup{width=0.8\textwidth}
    \caption{Accuracy vs. number of heads on different data sets}
    \label{fig:acc_num_heads}
\end{figure*}

\paragraph{Uncertainty measures comparison}
The uncertainty measure is designed to assess the quality of the model predictions. Failure prediction could be addressed by finding a good separator over a selected uncertainty measure, such as \cref{eqn2,eqn3,eqn4}. We average each of the uncertainty measures on the correct predictions and on the incorrect ones separately. A pronounced difference between the mean uncertainty measures of the true and false predictions might indicate the existence of a good separator. Based on the application needs, a threshold  on the uncertainty measure could be set, in such a case. The results are shown in \autoref{tbl:unct_comp}.
\begin{table}[ht!]
 \centering
  \caption{Uncertainty comparison}
  \label{tbl:unct_comp}
  \begin{tabular}{llcccccc}
    \toprule
     Arch/Data set &Method &\multicolumn{2}{c}{eqn2} & \multicolumn{2}{c}{eqn3}  & \multicolumn{2}{c}{eqn4}\\
      \multicolumn{1}{c}{} & & True &False &True &False &True &False \\
    \cmidrule(r){1-8}

            & MC-dropout         &0.043 &0.349 &0.033 &0.236 &0.006 &0.072 \\
            & DE     &0.009 &0.186 &0.005 &0.107 &0.003 &0.056 \\
VGG16/SVHN  & DSE &0.893 &0.904 &0.679 &0.709 &0.148 &0.129 \\
            & SWA                &0.004 &0.116 &NA &NA &NA &NA \\
            &Ours                &0.343 &0.544 &0.113 &0.247 &0.163 &0.156 \\

    \cmidrule(r){1-8}
                      &MC-dropout          &0.193 &0.486 &0.163 &0.406 &0.017 &0.044 \\
                      &DE      &0.058 &0.28 &0.033 &0.156 &0.016 &0.081  \\
MobileNet-v2/CIFAR10  &DSE  &0.092 &0.337 &0.079 &0.337 &0.006 &0.03  \\
                      &SWA                 &0.021 &0.16 &NA &NA &NA &NA              \\
                      &Ours                &0.11 &0.217 &0.0004 &0.002 &0.061 &0.061 \\

    \cmidrule(r){1-8}
                     & MC-dropout          &0.229 &0.514 &0.21 &0.495 &0.004 &0.009 \\
                     & DE      &0.276 &0.549 &0.237 &0.472 &0.024 &0.043 \\
ResNet50/CIFAR100    & DSE  &0.567 &0.673 &0.403 &0.517 &0.151 &0.103 \\
                     & SWA                 &0.207 &0.482 &NA &NA &NA &NA \\
                     &Ours &0.305 &0.545 &0.273 &0.514 &0.023 &0.015 \\

    \cmidrule(r){1-8}
                     & MC-dropout          &0.275            &0.563       &0.357    &0.448   &0.013   &0.036        \\
                     & DE      &0.357            &0.534       &0.364  &0.443    &0.026 &0.046         \\
Xception/ImageNet    & DSE  &0.757            &0.823       &0.433   &0.663  &0.132  &0.129           \\
                     & SWA                 &0.203            &0.452       &NA   &NA &NA    &NA           \\
                     &Ours                 &0.317            &0.499       &0.225    &0.431    &0.164  &0.157           \\
    \cmidrule(r){1-8}
  \end{tabular}
\end{table}
\paragraph{Accuracy vs. Number of heads}
To evaluate the effect of increasing the number of heads in our method, we train Mobile-NetV2 \cite{mobilenetv2} on CIFAR10 and VGG16 \cite{VGG} on SVHN with 1 to 4 heads and compare their obtained accuracy. The heads were added in the following order:
for CIFAR10, \cite{relaxed_SM}, \cite{sensoy2018evidential}, Mean Squared Error (MSE), Mean Absolute Error (MAE);
for SVHN, \cite{relaxed_SM}, \cite{sensoy2018evidential}, \cite{l_dmi_loss}, MSE.
The results are summarized in \autoref{fig:acc_num_heads}.
\section{Discussion}
\paragraph{Accuracy and calibration analysis}
As demonstrated in \autoref{tbl:basic_comp}, the proposed method improves accuracy on SVHN and CIFAR10 as well as ImageNet. On these data sets, the expected calibration error does not surpass each of the compared methods but remains competitive (ECE$\leq5\%$). Intuitively, the higher ECE could stem from the fact that the models, having trained their own features, may come to provide different \textit{opinions} on the same query image, in contrast to DSE using more similar models, or SWA using a single model. This conjecture yet remains to be rigorously assessed.
It is worth noting that the ECE is lower than the one reported in \cite{TS_MC_drop}, presumably due to the use of novel loss functions in our method \cite{sensoy2018evidential,relaxed_SM}. 
In regards to Brier score, our results are almost identical to the best methods we compare against (absolute diff. $\leq0.008$), as can be seen in \autoref{tbl:basic_comp}.
On CIFAR100, accuracy was improved over MC-dropout and DSE, though did not outperform DE and SWA. Deep Ensembles is a lot heavier framework than ours and SWA is known as one of the best known methods in the field.
\paragraph{Uncertainty measures}
Uncertainty is often measured by performing out-of-distribution analysis, by training a model on a subset of the classes considered in-distribution and testing on the remaining classes. We alternatively propose to use uncertainty measures for pointing out mispredictions even on in-distribution data. Intuitively, we should expect from a good uncertainty measure, provided a well-performing network, to exhibit a great difference between the sets of correct and incorrect predictions. The largest the difference is, the easier will it be to determine a threshold on the uncertainty measure, separating correct from incorrect predictions. The threshold is use-case dependent and should be set with regards to the allowed rate of false positives or false negatives.
\section{Conclusion}
In this work, we propose to adopt a multi-task approach for single-task variational inference and uncertainty estimation. We show that the approach improves accuracy over the most relevant variational inference methods on several data sets. Additionally, its relatively low run-time requirements make it feasible for practical use cases. Furthermore, we introduce an intuitive way to evaluate uncertainty estimation and to detect mispredictions, even on in-distribution data. 
For future work, more sophisticated forms of dropout such as \cite{ghiasi2018dropblock,chen2020dropcluster} could be used to perform variational inference. A regularizer to the different sub-ensembles could be introduced for enhanced control on the consistency of the different predictions, while preserving the gain in accuracy obtained by the diversity in the heads. 
It is also worth noting that \cite{corbiere2019addressing,jiang2018trust, ahuja2019probabilistic} could be combined with our method, i.e. training a Confid-Net branch for each network head, or using deep feature modeling based on the shared latent space.
Aiming at improving separations between true and false predictions might also be an interesting field for active research.
The uncertainty measures employed in this work could be used to evaluate the best uncertainty separator for detecting mispredictions, and by such, to bring the system to reach some level of awareness of its own limitations.
\small

\clearpage
\printbibliography
\end{document}